\documentclass[conference]{IEEEtran}
\IEEEoverridecommandlockouts
\usepackage{cite}
\usepackage{amsmath,amssymb,amsfonts}
\usepackage{graphicx}
\usepackage{textcomp}
\usepackage{xcolor}

\usepackage{multirow}
\usepackage[hidelinks]{hyperref}

\usepackage{float}

\usepackage{tikz, tikzscale}
\usetikzlibrary{shapes.geometric}
\usetikzlibrary{positioning}
\usepackage{pgfplots}

\usepackage{stackengine}

\usepackage{footnote}

\usepackage{algorithm}
\usepackage[noend]{algpseudocode}

\usepackage{float}
\floatstyle{plaintop}
\restylefloat{table}

\usepackage[justification=centering]{caption}

\usepackage{caption}
\usepackage{subcaption}

\def\BibTeX{{\rm B\kern-.05em{\sc i\kern-.025em b}\kern-.08em
    T\kern-.1667em\lower.7ex\hbox{E}\kern-.125emX}}
\begin{document}

\title{
       DoPose-6D dataset for object segmentation \\
       and 6D pose estimation
\thanks{
This work is funded by ML2R – The Competence Center Machine Learning Rhine-Ruhr which is funded by the Federal Ministry of Education and Research of Germany (grant no. 01|S18038A). \\
}
}

\author{
\IEEEauthorblockN{
Anas Gouda,
Abraham Ghanem,
Christopher Reining}
\IEEEauthorblockA{
Chair of Materials Handling and Warehousing, TU Dortmund University, Germany\\
\{anas.gouda,abraham,ghanem,christopher.reining\}@tu-dortmund.de
}
}

\maketitle

\begin{abstract}
Scene understanding is essential in determining how intelligent robotic grasping and manipulation could get. It is a problem that can be approached using different techniques: seen object segmentation, unseen object segmentation, or 6D pose estimation. These techniques can even be extended to multi-view. Most of the work on these problems depends on synthetic datasets due to the lack of real datasets that are big enough for training and merely use the available real datasets for evaluation.

This encourages us to introduce a new dataset (called DoPose-6D). The dataset contains annotations for 6D Pose estimation, object segmentation, and multi-view annotations, which serve all the pre-mentioned techniques. The dataset contains two types of scenes bin picking and tabletop, with the primary motive for this dataset collection being bin picking.

We illustrate the effect of this dataset in the context of unseen object segmentation and provide some insights on mixing synthetic and real data for the training. We train a Mask R-CNN model that is practical to be used in industry and robotic grasping applications. Finally, we show how our dataset boosted the performance of a Mask R-CNN model.

Our DoPose-6D dataset, trained network models, pipeline code, and ROS driver are available online
\footnote{
\url{https://github.com/AnasIbrahim/image_agnostic_segmentation}
\label{footnote:github}
}.

\end{abstract}


\section{Introduction}

\begin{figure}
    \centering
    \includegraphics[trim={300 0 300 150},clip,width=0.23\textwidth]{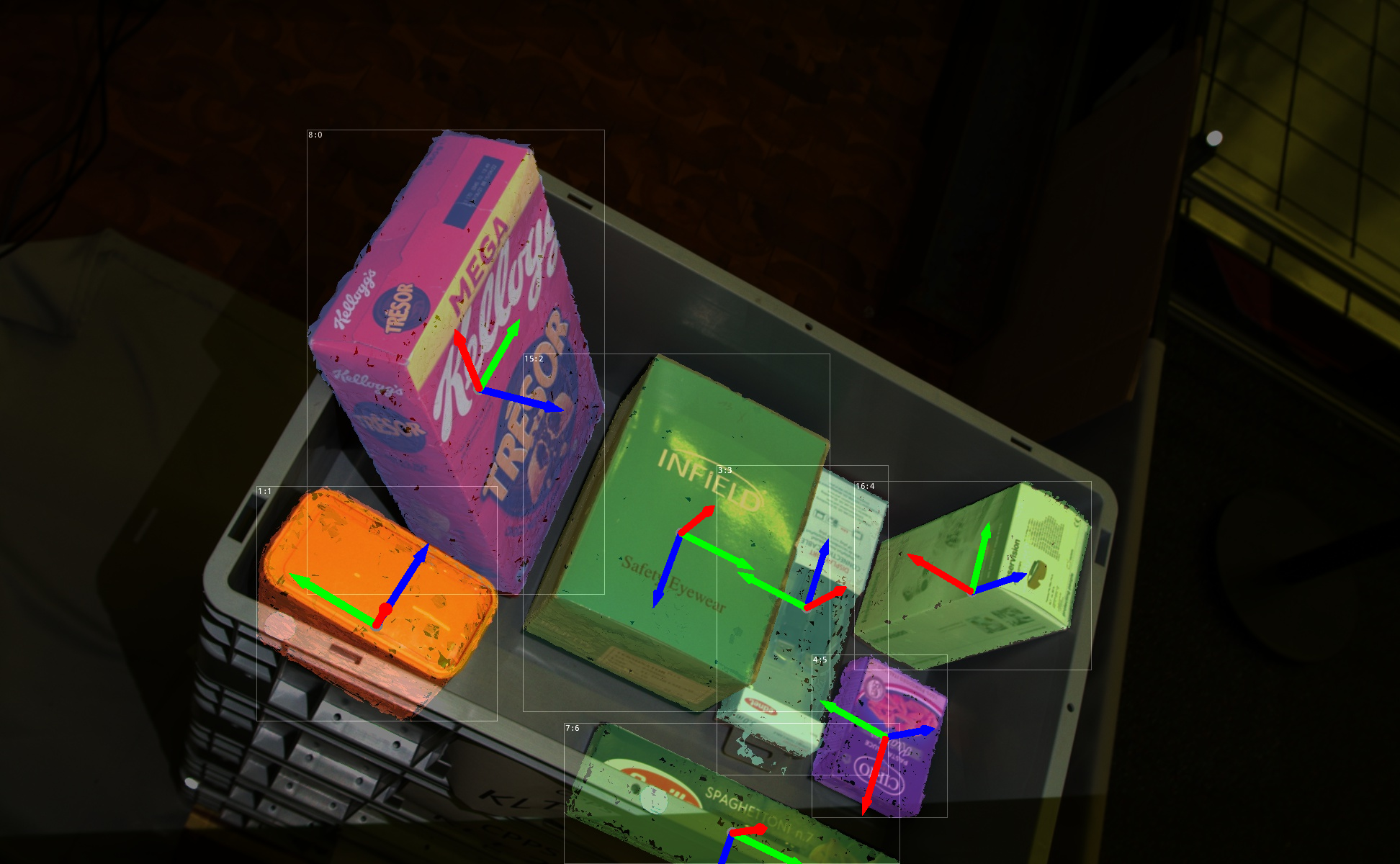}
    \includegraphics[trim={300 150 300 0},clip,width=0.23\textwidth]{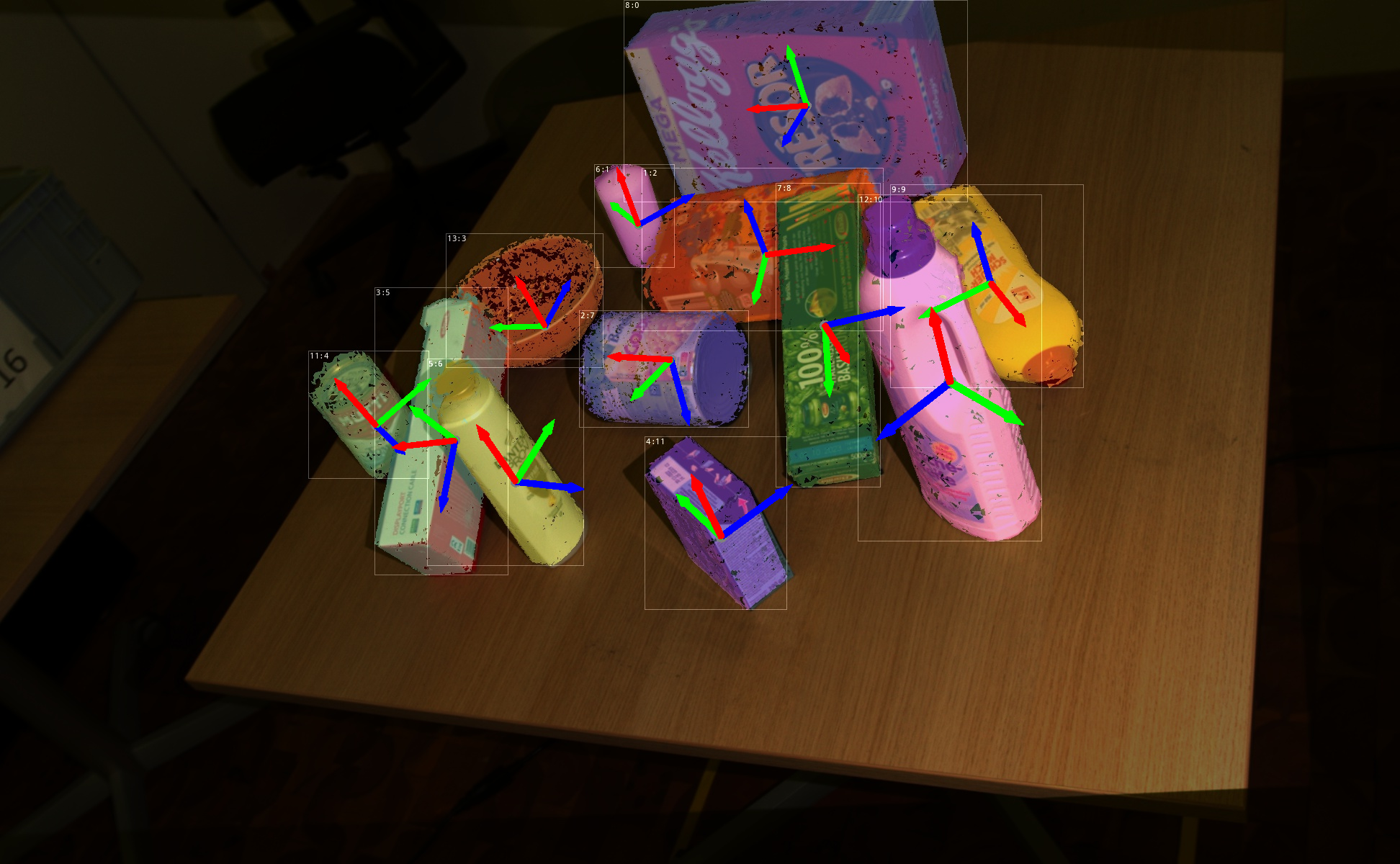}\\
    \vspace{1mm}
    \includegraphics[trim={300 150 300 0},clip,width=0.23\textwidth]{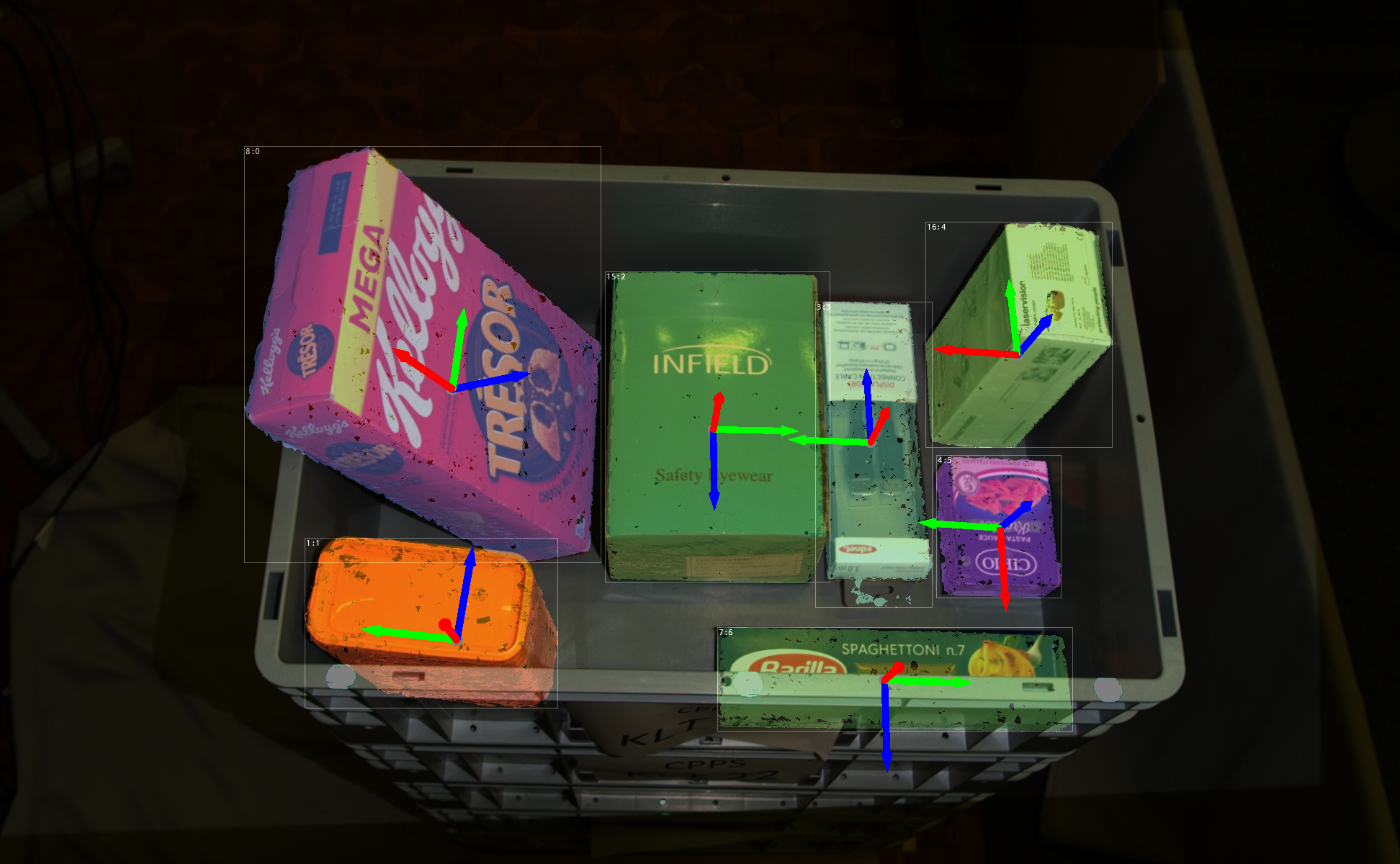}
    \includegraphics[trim={300 150 300 0},clip,width=0.23\textwidth]{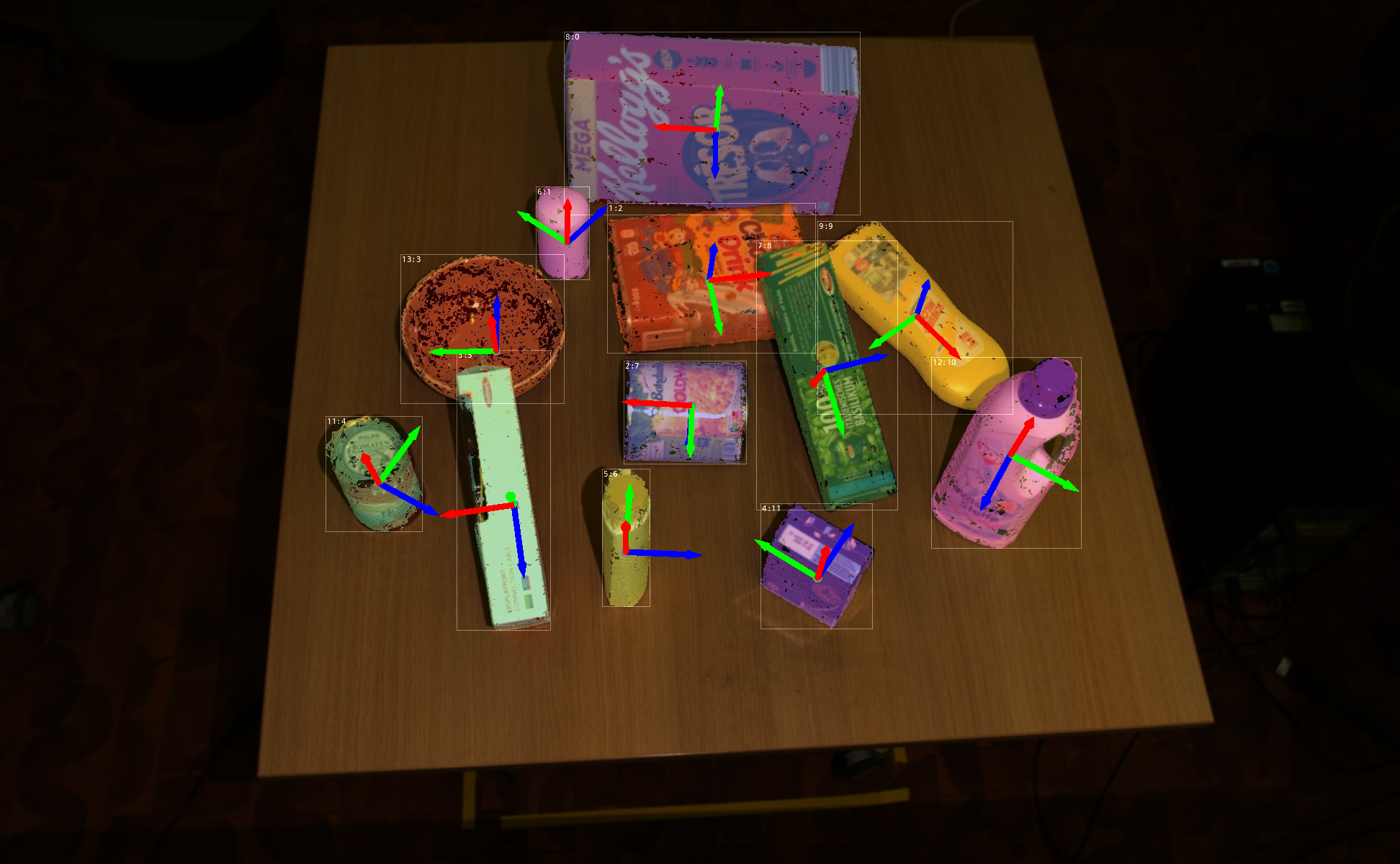}\\
    \vspace{1mm}
    \includegraphics[trim={50 125 550 25},clip,width=0.23\textwidth]{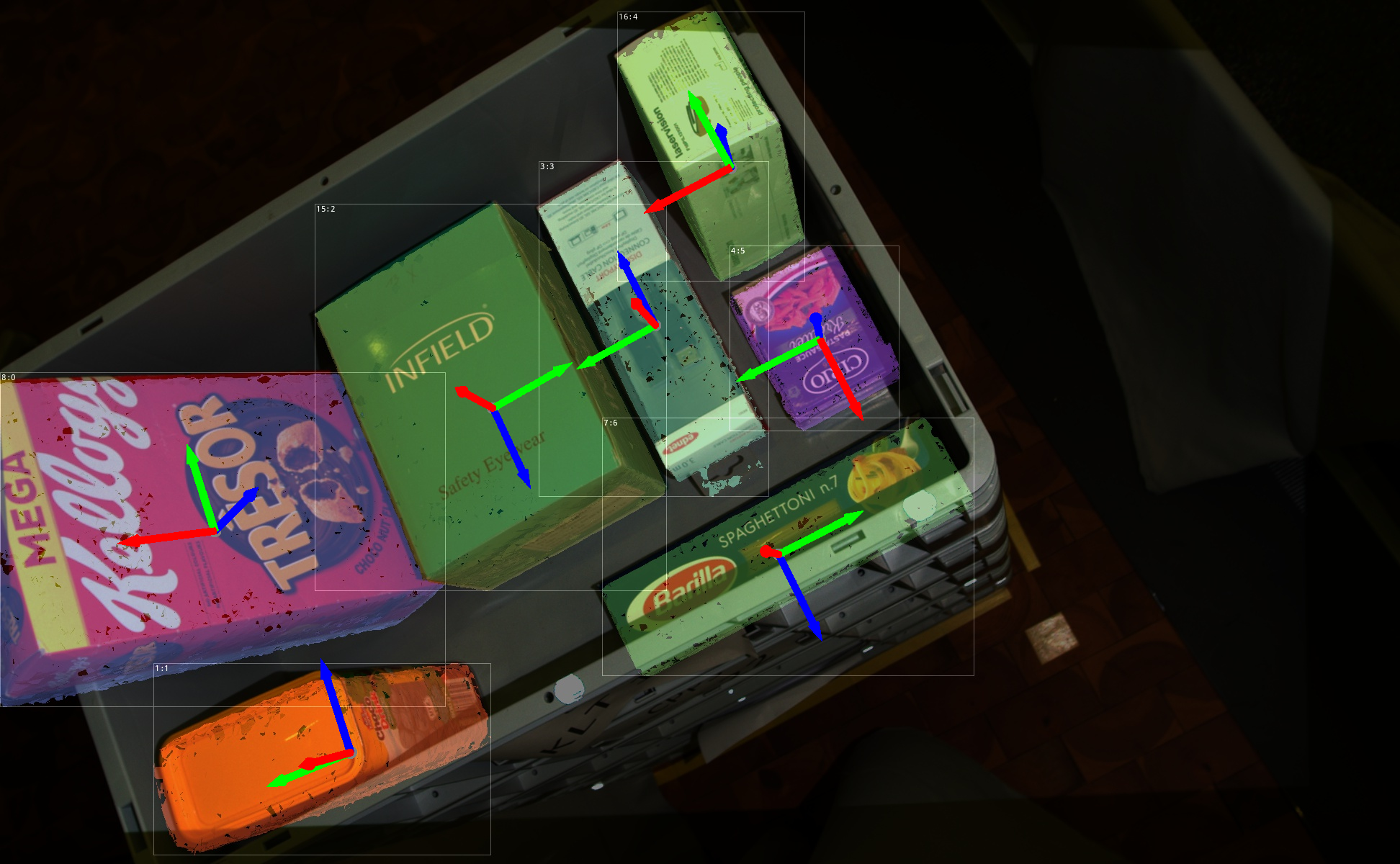}
    \includegraphics[trim={300 125 300  25},clip,width=0.23\textwidth]{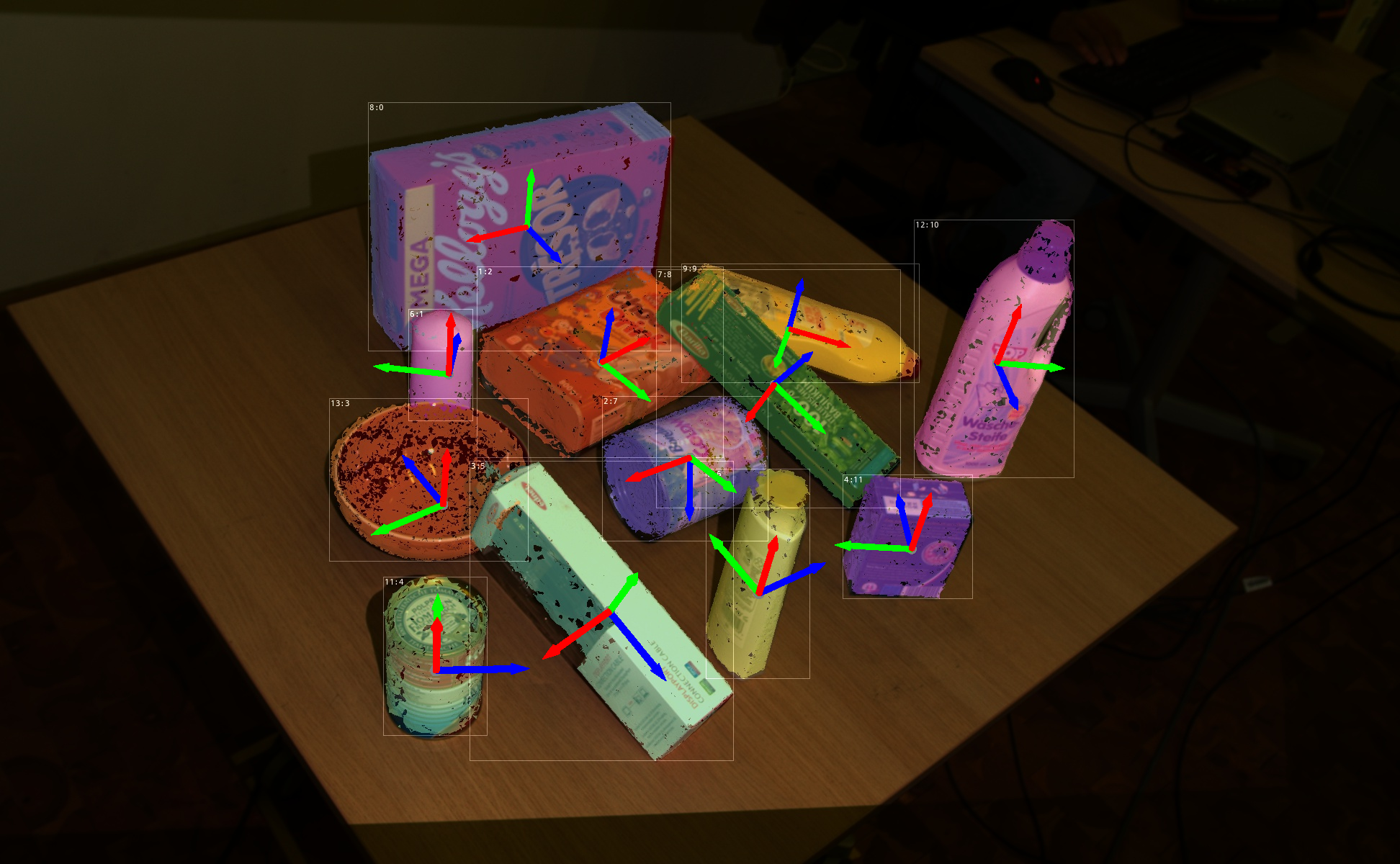}\\
    \caption{Two different samples of our DoPose-6D dataset for bin picking and tabletop. Each sample contains multi-views of a scene. Samples contain between 2 to 16 views. 6D annotations were manually made for the center view then projected to other views using the scene transformations.
    }
    \label{fig:view_angles}
\end{figure}

Dataset availability drives the possibility of the deep learning methods' existence. In the case of domains where real datasets do not exist, researchers opt for synthetic datasets because creating a real dataset is impossible in an economically feasible and timely manner. 
This generates the sim2real gap. While considerable research showed that training on synthetic data can still generalize well to real data, having a dataset to retrain or fine-tune CNN models for real-world usage is still crucial. Using a real dataset for fine-tuning enhances the performance of models and makes them more reliable. Robotic grasping and specifically bin picking are one of those applications where real datasets shortage exists. This motivates us to create a real dataset for robotic grasping and bin picking.

The reason why a dataset is specifically made for bin picking is that they exhibit more clutter and higher occlusion with constrained grasping poses. A non-accurate segmentation mask can lead a suction gripper to compute a grasping point on an object corner and leads the object to drop while moving it after the grasp. Also, Using bins as a storage unit is becoming more common, and warehouse storage systems such as AutoStore are becoming widely adopted. However, most of these bin-based warehouses still depend on manual picking for reliability, as most automated bin-picking cells still rely on classical algorithms for segmentation and cannot handle a wide range of objects. This drives us to the concept of category-agnostic and unseen object segmentation. A CNN that is robustly capable of segmenting different types of unseen objects would lead to full automation of such picking cells. Also, as shown later in section \ref{sec:results}, Fine-tuning on bin picking scenes would enhance the performance of our trained model on other scenes such as tabletop.

Section~\ref{sec:related_work} justify the creation of our dataset with unseen object segmentation in mind. Section~\ref{sec:method} explains the collection process for our DoPose-6D real dataset and explains how a synthetic dataset and our dataset are used to train a model for unseen object segmentation. Finally in section~\ref{sec:results}, we evaluate our model on unseen datasets, show how our DoPose-6D dataset affected the models performance, compare the model performance to other methods and elaborate how our model is used to compute suction grasp poses.




The key points of this contribution consist of:
\begin{itemize}
    \item A dataset of highly cluttered and closely stacked objects on two different scene types (tabletop, bin picking) with HDR (high dynamic range) images. The dataset offers the largest collection of real images of bin scenes with annotations for several segmentation and object detection tasks.
    \item A model for unseen object segmentation that is trained and evaluated on different scene environments.
    \item ROS package for unseen object segmentation and suction grippers based on our trained models.
\end{itemize}
\section{Related Work}
\label{sec:related_work}

\begin{table}[]
    \centering
    \resizebox{\columnwidth}{!}{
    \begin{tabular}{|c|c|c|c|c|}
        \hline
        Name                                     & Environment      & Objects & Scenes & Samples \\
        \hline
        \hline
        DoPose-6D                                   & tabletop \& bin  &   18    & 301           & 3325 \\
        \hline
        HOPE \cite{tyree2022hope}                &   various        &   28    & 50            & 238 \\
        \hline
        YCB-V \cite{xiang2018posecnn}            &   various        &   21    & 92            & 133827 \\
        \hline
        LM \cite{10.1007/978-3-642-37331-2_42}   &   tabletop       &   15    & 15            & 18273 \\
        \hline
        LM-O \cite{10.1007/978-3-319-10605-2_35} &   tabletop       &   8     & \textless 25  & 1214 \\
        \hline
        HB \cite{kaskman2019homebreweddb}        &   tabletop       &   33    &  13           & 17420 \\
        \hline
        T-LESS \cite{hodan2017tless}             &   tabletop       &   30     & 20           & 10080 \\
        \hline
        ITODD \cite{8265467}                     &   top-down view  &   28     & 776          & 766 \\
        \hline
        
    \end{tabular}
    \caption{Comparing our DoPose-6D dataset to real datasets included in BOP challenge\cite{hodavn2020bop}. Scenes is the number of individual object arrangement. Samples is the total number of view images for all scenes.}
    \label{tab:datasets_comparison}
    }
\end{table}

Several datasets for object segmentation and object pose estimation exist. These datasets can be categorized into one of three: real dataset, synthetic dataset, or a data generator.

YCB-V \cite{xiang2018posecnn}, LM \cite{10.1007/978-3-642-37331-2_42}, LM-O \cite{10.1007/978-3-319-10605-2_35}, T-LESS \cite{hodan2017tless}, ITODD \cite{8265467}, and HOPE \cite{tyree2022hope} datasets provided 6D pose annotations either for tabletop or various cluttered objects scenes. BlenderProc generator \cite{denninger2019blenderproc} was used during the BOP challenge \cite{hodavn2020bop} to generate photo-realistic synthetic data for these datasets. Table~\ref{tab:datasets_comparison} shows a comparison between our dataset and these real datasets included in the BOP challenge \cite{hodavn2020bop}. According to this comparison the main plus point for our DoPose-6D dataset is the focus on bin picking scenes. Also our dataset covers the highest number of unique scenes which provides high variation for the training.

Stillleben generator \cite{9197309} was used to generate SynPick dataset \cite{9551599} also with 6D pose annotations. NVIDIA Dataset Synthesizer \cite{to2018ndds} was used to generate NVIDIA FAT dataset \cite{DBLP:journals/corr/abs-1804-06534}. Several other works used PyBullet engine \cite{coumans2021} to generate synthetic data with lower quality rendering process.

OCID \cite{8793917} and OSD \cite{6385661} provided 2D segmentation for tabletop \& floor scenes and WISDOM \cite{danielczuk2019segmenting} for bin scenes. WISDOM dataset is the only dataset that includes bin picking scenes but only included $300$ images with a fixed top-down camera pose and 2D segmentation only.


The problem of category-agnostic segmentation of unseen objects for robotics grasping was discussed in SD Mask R-CNN \cite{danielczuk2019segmenting}. Their network used a divergent of Mask R-CNN \cite{he2017mask} to segment unseen objects using depth images, with the main environment for segmentation being bin picking. \cite{9680821} proposed a pipeline to split object segmentation and classification for unseen object into two steps. UCN \cite{xiang2020learning} used RGB-D feature embeddings followed by a clustering algorithm and then a refinement step. UOIS \cite{xie2021unseen} introduced a network to segment unseen objects using 3D data representation instead of RGB or RGB-D images. The same authors in RICE \cite{xie2021rice} refined segmentation masks using a graph based representation to separate under-segmented and join over-segmented objects. RICE enhanced the performance of several of the pre-mentioned methods.

Even though many other datasets exist, they are either synthetic, only provide a small number of individual scenes with small variation in individual object configurations, or provide tabletop scene setups only. 
This encouraged us to collect and publish our novel DoPose-6D dataset. It covers highly occluded closely stacked objects for 2 scene setups, mainly bin picking and also tabletop. These different setups along the high number of individual samples will cover sufficient data variance so that our models generalize better on unseen data, as discussed in section \ref{sec:results}. The dataset contains a various set of annotations that is suitable for different segmentation and detection tasks for 2D and 3D methods.

\definecolor{bblue}{HTML}{4F81BD}
\definecolor{rred}{HTML}{C0504D}
\definecolor{ggreen}{HTML}{9BBB59}
\definecolor{ppurple}{HTML}{9F4C7C}


\begin{figure}
    \centering
    \begin{tikzpicture}
        \begin{axis}[
            symbolic x coords={
                pantene shampoo,
                HDMI cable,
                corn can,
                cereal box,
                tomato can,
                scheuermilch,
                waschesteife,
                safety glassbox,
                barilla spaghetti,
                krauter sauce,
                infield box,
                coca cola bottle,
                choco box,
                red bowl,
                scissors,
                stan southpark,
                white candle,
                heinz ketchup
            },
            xtick=data,
            ylabel=object count,
            xticklabel style={rotate=45,anchor=north east},
            xbar legend=fill,
            style={font=\footnotesize},
          ]

            \addplot[ybar,fill=ggreen, opacity=1.0] coordinates {
                (choco box, 350)
                (corn can, 320) 
                (HDMI cable, 314)
                (krauter sauce, 945)
                (pantene shampoo, 68)
                (white candle, 63)
                (barilla spaghetti, 902)
                (cereal box, 331)
                (scheuermilch, 380) 
                (scissors, 192)
                (tomato can, 358)
                (waschesteife, 554)
                (red bowl, 239)
                (heinz ketchup, 48)
                (infield box, 697)
                (safety glassbox, 736)
                (stan southpark, 62)
                (coca cola bottle, 640)
            };

            \addplot[ybar,fill=rred, opacity=0.7] coordinates {
                (choco box, 227)
                (corn can, 314)
                (HDMI cable, 295)
                (krauter sauce, 469)
                (pantene shampoo, 266)
                (white candle, 156)
                (barilla spaghetti, 403)
                (cereal box, 298)
                (scheuermilch, 357)
                (scissors, 204)
                (tomato can, 250)
                (waschesteife, 335)
                (red bowl, 264)
                (heinz ketchup, 102)
                (infield box, 238)
                (safety glassbox, 170)
                (stan southpark, 136)
                (coca cola bottle, 20) 
            };
            
            \legend{bin scenes, tabletop scenes}
        \end{axis}
    \end{tikzpicture}
    \caption{DoPose-6D dataset object distribution}
    \label{fig:dataset_objects_dist}
\end{figure}
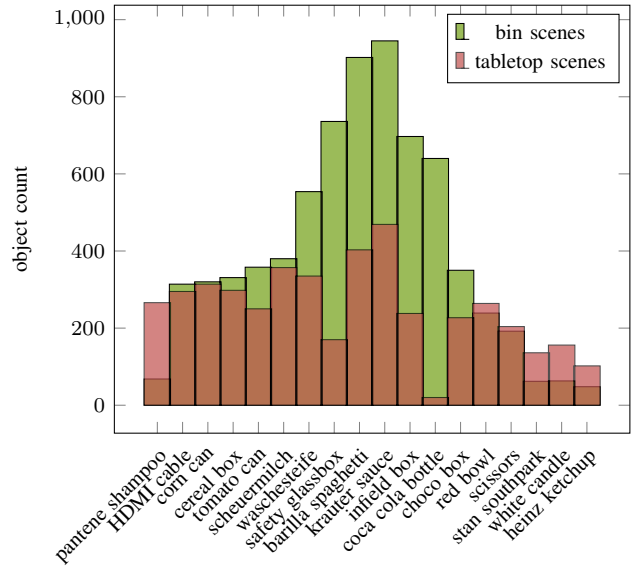

\begin{figure*}
    \centering
    \begin{subfigure}[b]{0.42\textwidth}
        \begin{tikzpicture}
          \begin{axis}[
              width=\linewidth, 
              grid=major, 
              grid style={dashed,gray!30}, 
              xlabel=Epochs ,
              ylabel=Loss , 
              legend pos=north east, 
              ymin=0, ymax=1.2
            ]
            \addplot
            [color=black,opacity=0.6, very thick]
            table[x=Step,y=Value,col sep=comma, smooth, tension = 0.1, each nth point = {10},
            x expr=\thisrowno{1}*8/(31500*0.97)]
            {./training_validation_data/training_loss_split.csv};
            \addplot 
            [color=blue,opacity=0.6, very thick]
            table[x=Step,y=Value,col sep=comma,
            x expr=\thisrowno{1}*8/(31500*0.97)] {./training_validation_data/validation_loss_split.csv};

            \legend{
            training loss,
            validation loss}
          \end{axis}
        \end{tikzpicture}
        \caption{Splitting a dataset between training and validation}
    \end{subfigure}
    \begin{subfigure}[b]{0.42\textwidth}
        \begin{tikzpicture}
          \begin{axis}[
              width=\linewidth, 
              grid=major, 
              grid style={dashed,gray!30}, 
              xlabel=Epochs ,
              ylabel=Loss , 
              legend pos=north east, 
              ymin=0, ymax=1.2
            ]
            \addplot
            [color=black,opacity=0.6, very thick]
            table[x=Step,y=Value,col sep=comma, smooth, tension = 0.1, each nth point = {10},
            x expr=\thisrowno{1}*8/(31500*0.97)]
            {./training_validation_data/training_loss_different_datasets.csv};
            \addplot 
            [color=blue,opacity=0.6, very thick]
            table[x=Step,y=Value,col sep=comma,
            x expr=\thisrowno{1}*8/(31500*0.97)]
            {./training_validation_data/validation_loss_different_datasets.csv};

            \addplot[red] coordinates {(0.65,0.1) (0.65,0.9)};
            \addplot[red] coordinates {(0.1,0.49) (0.9,0.49)};
            \legend{
            training loss,
            validation loss}
          \end{axis}
        \end{tikzpicture}
        \caption{different datasets for training and validation}
    \end{subfigure}
    \caption{An essential concept when training a model for unseen object segmentation is not to split the training dataset between training and validation. Splitting the dataset between training and validation would lead to a misleading, forever-decreasing validation loss, as shown in (a). This is because when a dataset is split, the validation data is similar to the training data (exact same objects). Instead, a different dataset should be used for validation (different object categories). In the case of our first training step, NVIDIA FAT dataset \cite{DBLP:journals/corr/abs-1804-06534} was used for training, and our DoPose-6D dataset was used for validation. As shown in (b), the validation loss increases after 0.65 epochs.}
    \label{fig:training_loss}
\end{figure*}
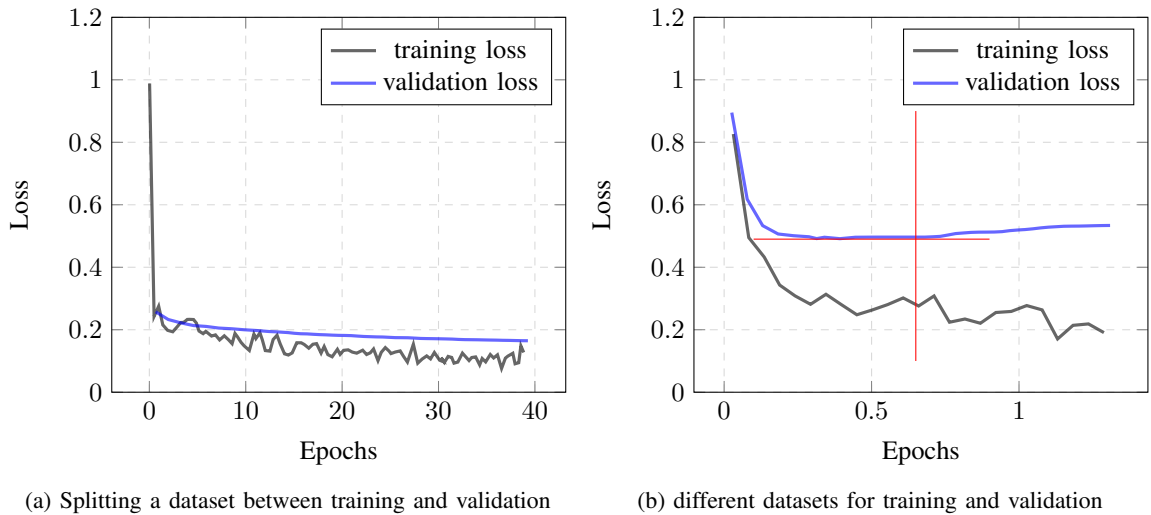

\section{Method}
\label{sec:method}

In this section, we first point out the data collection process of our DoPose-6D dataset. Second, we explain how the first training step is carried out using synthetic data and the second training step using real data.


\subsection{Dataset and annotation tool}
\label{sec:dataset_creation}

As manual annotation is time-consuming and labor intensive, our data collection process is made semi-automated. We use a Zivid Two structured light camera mounted on a Kuka iiwa LBR 14 robotic manipulator. For each scene, RGB images, depth images, and the camera transformations of different view angles are captured as shown in figure~\ref{fig:view_angles}. The dataset is saved in the BOP format \cite{hodan2018bop}, a standardized format for 6D pose estimation. Using this format facilitates our dataset creation process.

For each of our objects, we created a 3D mesh model. These 3D models were made by capturing all surfaces of each object by a Zivid Two camera and assembling them using the CloudCompare software. The dataset contains 18 object covering different shapes, sizes, colors, and textures.

We use our annotation tool to manually annotate the 6D pose of each object of the centered sample (Figure \ref{fig:view_angles} middle image) with the created 3D meshes. The annotation tool using the camera transformation calculates the 6D pose of all objects in all view angles. The BOP toolkit is then used to generate the segmentation masks for all samples and angles along with the COCO format JSON files.

We include one extra annotation file that is not standardized in the BOP format: the scene transformations between the camera for each view angle and the world frame. We also accompany our code with a script that generates annotated point cloud data. This way, our dataset can be used for multiple purposes: Instance segmentation, 6D Pose estimation, and point cloud segmentation along with multi-view methods.



For the bin picking samples, the data contains 183 scenes with 2150 image views. Of those 183 scenes, 35 scenes contain 2 views, 20 scenes contain 3 views, 28 scenes contain 15 views, and 100 scenes contain 16 views. For the tabletop samples, the data contains 118 scenes with 1175 image views. Of those 118 scenes, 20 scenes contain 3 views, 1 scene contains 5 views, 49 scenes contain 6 views, and 48 scenes contain 17 views. So in total, our data contains 301 scenes and 3325 view images. Most of the scenes contain mixed objects. The dataset includes 18 objects in total. 5 among these 18 objects were chosen to look a bit similar (not the same object) to YCB objects on purpose to be able to study the effect of generalization on similar and totally different objects as several existing datasets use the YCB objects. Figure \ref{fig:dataset_objects_dist} shows the distribution of the objects in the dataset.



\begin{figure*}
    \centering
    \begin{subfigure}[b]{0.42\textwidth}
    \includegraphics[trim={180 0 20 0},clip,width=\textwidth,keepaspectratio]
    {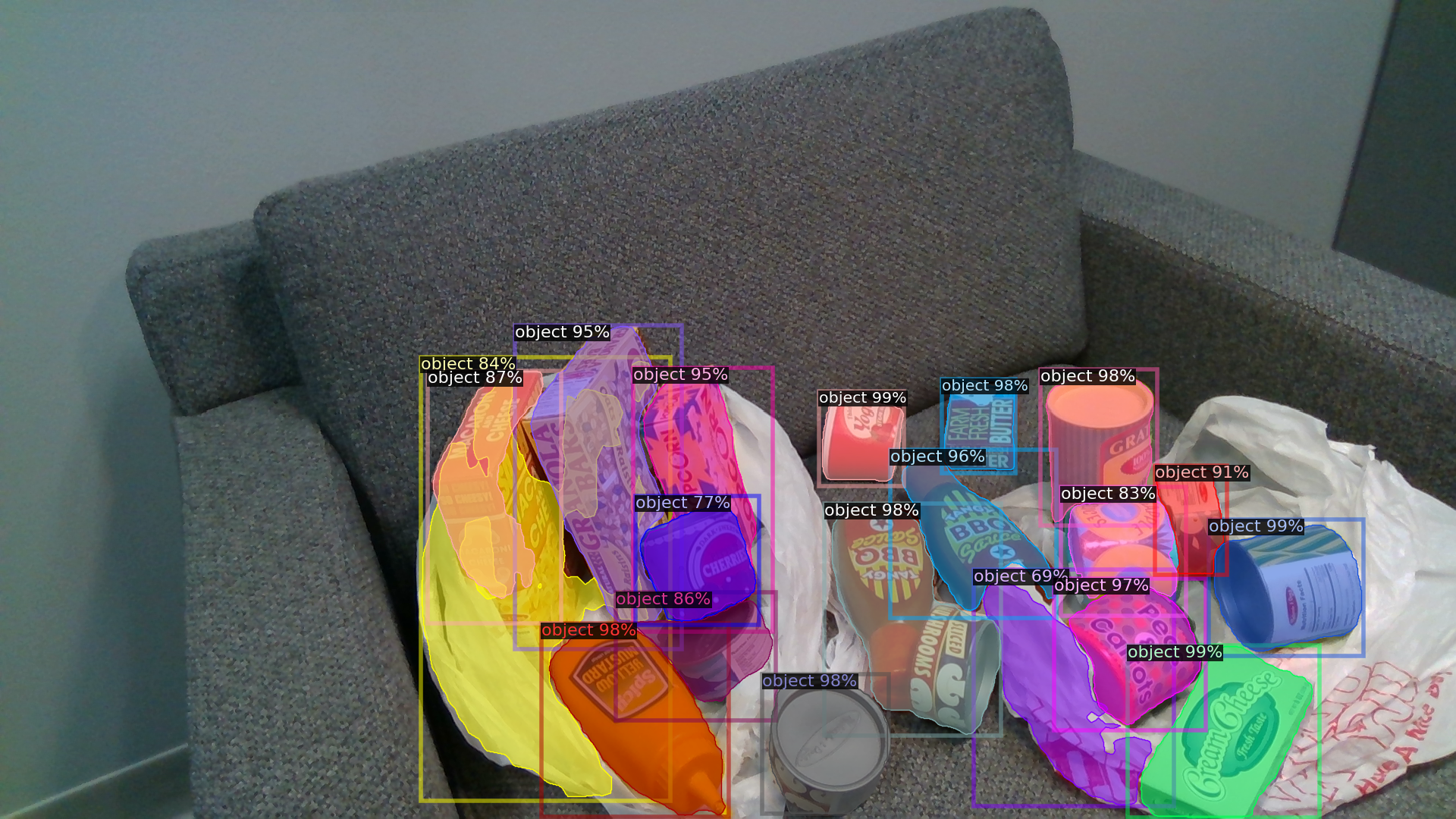}
    \caption{model trained on synthetic data NVIDIA FAT dataset}
    \end{subfigure}
    \begin{subfigure}[b]{0.42\textwidth}
    \includegraphics[trim={180 0 20 0},clip,width=\textwidth,keepaspectratio]
    {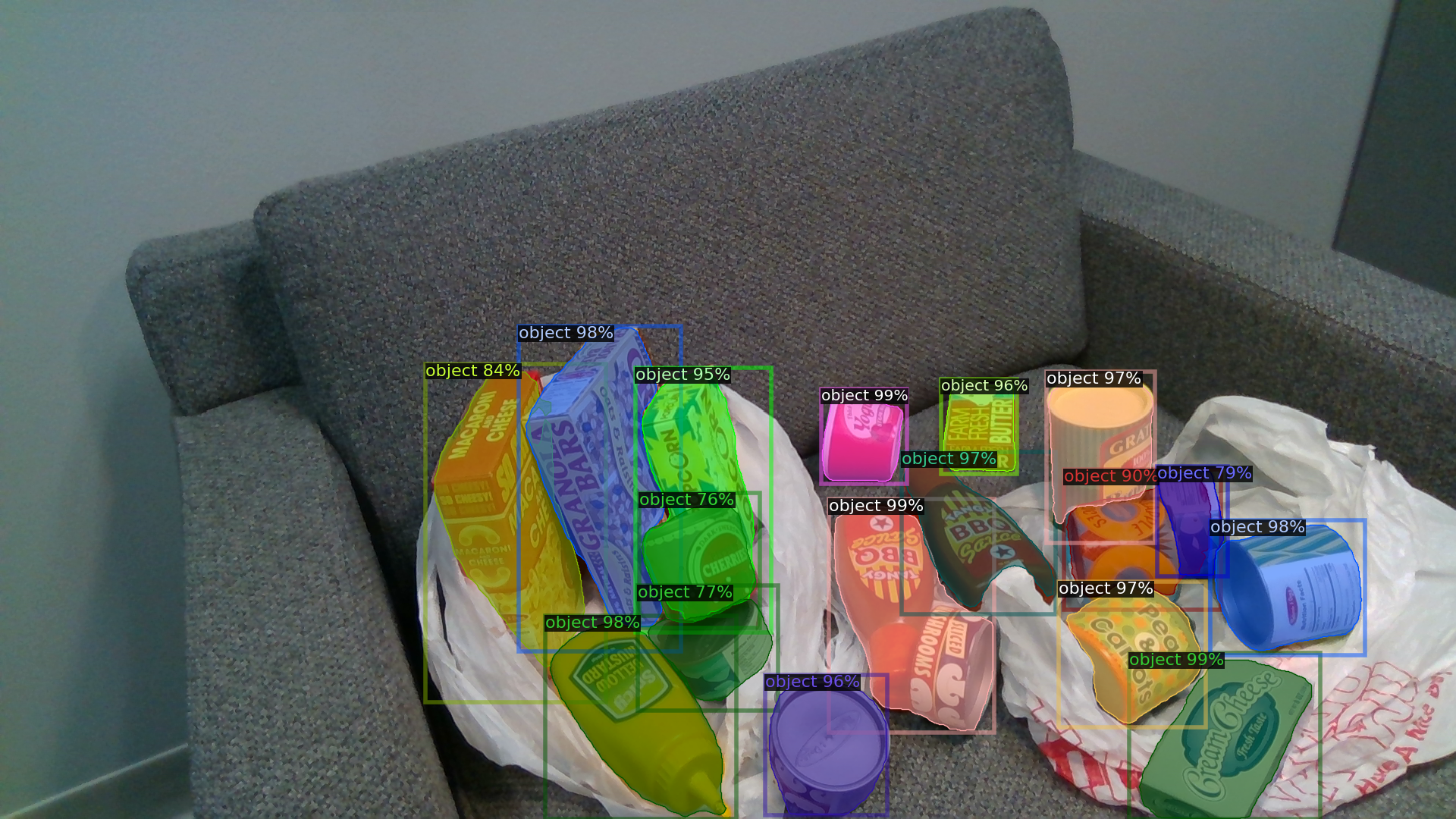}
    \caption{fine-tuned model on DoPose-6D}
    \end{subfigure}
    \caption{Result of segmentation on an unseen dataset with unseen objects from NVIDIA HOPE dataset \cite{tyree2022hope}. The figure shows the effect of fine-tuning the model using our DoPose-6D dataset. The effect is evident for closely stacked objects correctly segmented and removing random batches segmented from the background.}
    \label{fig:hope_dataset_segmentation_example}
\end{figure*}

\subsection{unseen object segmentation}
\label{sec:model_training}



The main point to keep in mind when training models for unseen object segmentation is avoiding over-fitting. This is important because over-fitting can bound the learned object categories to similar categories as the training dataset. Figure~\ref{fig:training_loss} (a) shows how training loss can easily be misleading when splitting a dataset between training and validation because the validation data then is very similar to the training data (same object categories). In the case of the usual instance segmentation with the COCO dataset, validation images contain different instance of the class (ex. differnet humans or a different set of cups and not the exact similar object). However, this is not the case when training with a limited number of specific objects. That means we require different object categories for the validation set to be able to train a category agnostic model. Therefore we use a different dataset (with different object and object categories) for the validation set. This is shown in figure~\ref{fig:training_loss} (b), where the validation loss started to increase after 0.65 epochs. This means the training should stop after around 0.6 epochs, according to the early stopping rule.

We use Mask R-CNN \cite{he2017mask} model implementation from detectron2 \cite{wu2019detectron2} training only with RGB data. The model is pretrained on COCO dataset \cite{10.1007/978-3-319-10602-1_48}. Then we carry out two training steps. The first training step uses the NVIDIA FAT \cite{DBLP:journals/corr/abs-1804-06534} synthetic dataset as the training dataset and DoPose-6D as the validation dataset for 0.6 epochs (figure~\ref{fig:training_loss} b). The second training step uses DoPose-6D as the training dataset and NVIDIA HOPE dataset \cite{tyree2022hope} for validation for 1.8 epochs.






\section{Results}
\label{sec:results}

In this section, we evaluate our model on unseen datasets with unseen objects in different environment setups. We compare our model to SOTA methods trained on different datasets. Finally, we elaborate how to compute suction grasps from the segmentation masks.

\subsection{Evaluation on unseen datasets}
\label{sec:unseen_eval}

\begin{figure}
    \centering
    \includegraphics[width=0.23\textwidth]{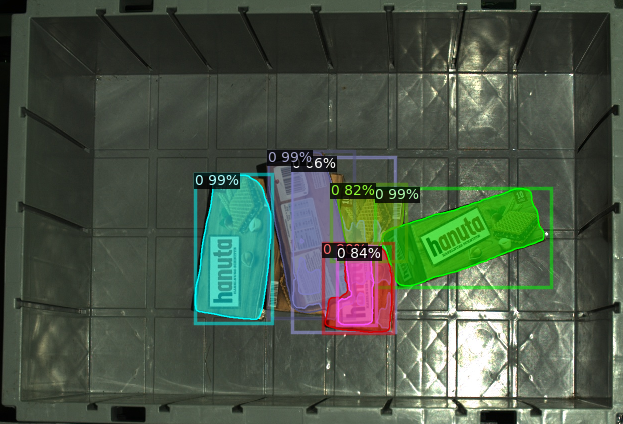}
    \includegraphics[width=0.23\textwidth]{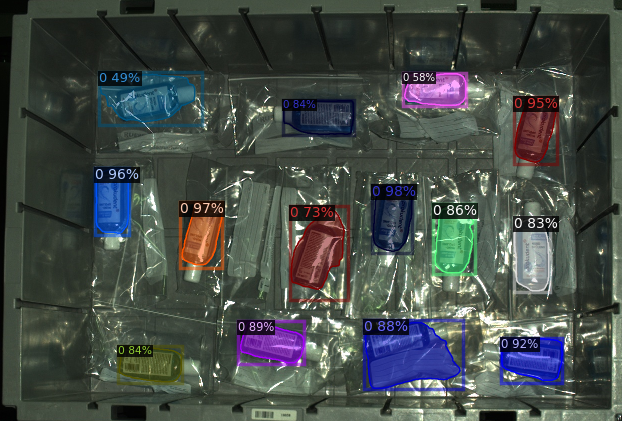}
    \caption{Segmentated samples of the proprietary dataset.}
    \label{fig:swisslog_example}
\end{figure}

\begin{table}
    \centering
    \caption{Evaluation on HOPE dataset and an industrial proprietary bin picking dataset}
    \begin{tabular}{|c| c|c| c|c|}
        \hline
        Dataset & \multicolumn{2}{|c|}{segmentation} & \multicolumn{2}{c|}{bounding box}\\
        \hline
        & AP & AR & AP & AR \\
        \hline
        \hline
        HOPE dataset            & 56.0 & 63.4 & 54.1 & 64.9 \\
        \hline
        Proprietary bin picking & 51.5 & 57.2 & 49.6 & 55.3 \\
        \hline
    \end{tabular}
    \label{table:unseen_datasets_evaluation}
\end{table}

We evaluate our model on the NVIDIA HOPE benchmark dataset \cite{tyree2022hope} which includes unseen objects on many random setups (tabletop, floor, shelf, couch, ..). Figure~\ref{fig:hope_dataset_segmentation_example} shows a sample from NVIDIA HOPE dataset before and after fine-tuning the model with our DoPose-6D dataset. The fine-tuning enhanced the model's performance for highly stacked objects and removed random batches segmented from the background. This is why we chose only to fine-tune the model using the bin scenes from our DoPose-6D dataset, as the bin scenes contain many samples of closely stacked and highly occluded objects.

We also evaluate our model on a proprietary bin-picking dataset from the industry. Figure~\ref{fig:swisslog_example} shows two segmented samples of the proprietary dataset. What makes our model practical for industrial bin picking is that the segmentation masks confidences are associated with occlusion and over/under-segmentation. So high confidence masks represent the objects that are correctly segmented and easier to grasp. On the contrary, lower confidence represents occluded objects or over/under-segmented objects. So a simple grasping approach would aim for segmentation masks with the highest confidence.

From figure~\ref{fig:hope_dataset_segmentation_example} and \ref{fig:swisslog_example}, we show that the same model was able to generalize on different environment setups. Table~\ref{table:unseen_datasets_evaluation} shows the COCO evaluation metrics for both dataset.

\subsection{Comparing to state-of-the-art}
\label{sec:sota_compare}

The main point of this section is to show how effective our DoPose-6D dataset contributed to enhancing the performance of the Mask R-CNN model and how well that model scored against other pre-trained approaches. The evaluation in this section is carried out using overlap P/R/F metrics (precision/recall/F-measure) introduced in \cite{dave2019towards} as used in \cite{xie2021unseen} which is computed by:

\vspace{3mm}

\resizebox{0.48\textwidth}{!}
{
\begin{math}
P = \frac{\sum_{i} |s_{i} \cap g(s_{i})|} { \sum_{i} |s_{i}| }
, R = \frac{\sum_{i} |s_{i} \cap g(s_{i})|} { \sum_{j} |g_{j}| }
, F = \frac{2PR}{P+R'}
\end{math}
}

\vspace{3mm}

We compare our model to two SOTA approaches. The first approach is UCN \cite{xiang2020learning} which extracts feature embeddings from RGB-D images and then uses a mean-shift clustering to cluster the features of unseen objects. The second method UOIS \cite{xie2021unseen}, leverages depth and RGB separately by computing object seeds from the depth and then using the seeds to segment the RGB image, followed by a final refinement step. Figure~\ref{fig:feature_embedding_sota} and \ref{fig:uois_sota} shows a comparison between our model and the two approaches. RICE \cite{xie2021rice} is a refining method that uses the segmentation output of any approach to join/split under/over-segmented masks.

The two approaches were evaluated in \cite{xie2021rice} which provides a comprehensive evaluation for unseen object segmentation methods. The evaluation for these methods, including ours, is carried out using the OCID dataset \cite{8793917}. Table~\ref{tab:sota_comapre} shows a comparison between the other methods and our model. This comparison is not a baseline comparison where all methods are trained on the same dataset. This is a comparison between other methods trained using TOD dataset \cite{pmlr-v100-xie20b} and our Mask R-CNN model fine-tuned on our DoPose-6D dataset.

The table shows that the same Mask R-CNN model performance was boosted from 80.3\% to 90.4\% reaching only 2.1\% less than UCN+RICE SOTA method. However, our Mask R-CNN model still scored less than Mask R-CNN + RICE. This shows that our DoPose-6D dataset effectively increased the core method's performance before refining.

\begin{table}
    \centering
    \begin{tabular}{|c|c|c|c|}
        \hline
        Method                        &  P   &   R  &   F  \\
        \hline
        
        Mask R-CNN		              & 80.3 & 79.8 & 79.3 \\
        Mask R-CNN + RICE	          & 92.3 & 91.8 & 92.0 \\
        

        \hline
        UOIS-Net-3D		              & 86.3 & 88.6 & 87.3 \\
        UOIS-Net-3D+RICE	          & 89.7 & 91.9 & 90.7 \\
        
        \hline
        UCN			                  & 91.6 & 92.8 & 91.9 \\
        UCN+RICE		              & 92.5 & 93.2 & 92.5 \\
        
        \hline
        ours (Mask R-CNN)              & 90.4 & 89.3 & 89.8 \\

        \hline
    \end{tabular}
    \caption{Comparison between our Mask R-CNN model fine-tuned on our DoPose-6D dataset, and other SOTA methods as evaluated in \cite{xie2021rice}. This is not a baseline comparison.}
    \label{tab:sota_comapre}
\end{table}

\subsection{Suction Grasp computation}

\begin{figure}
    \centering
    \includegraphics[trim={180 0 20 0},clip,width=0.48\textwidth,keepaspectratio]
    {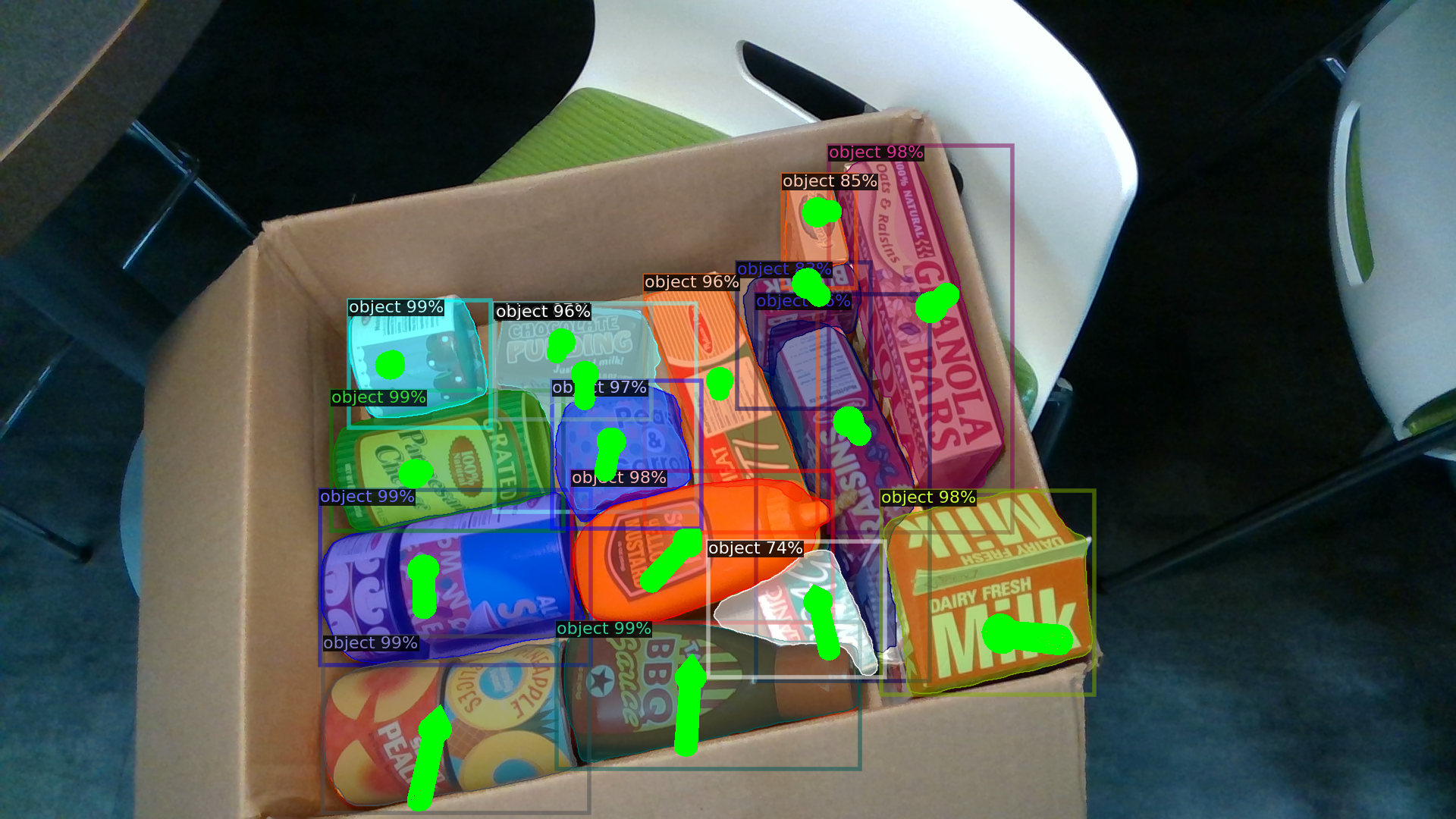}
    \caption{Computed suction grasp on a sample from NVIDIA HOPE dataset. Grasp poses are directed outward the camera}
    \label{fig:grasp_rendered}
\end{figure}

In this section, we show how to compute suction grasp points using our model. Algorithm \ref{alg:grasp} explains the step of calculating the grasp position and orientation. The position is the center of the biggest plane calculated by RANSAC. The orientation is the halfway vector of the normal at the center of this biggest plane. The orientation uses the halfway vector fixing the Y-axis (as the normal is a vector, not an orientation). Figure \ref{fig:grasp_rendered} shows the grasps computed for a sample from NVIDIA HOPE dataset.





\begin{algorithm}
\caption{Computing suction grasp point from a mask segment}
\label{alg:grasp}
\hspace*{\algorithmicindent} \textbf{Input: rgb depth matrix masks} \\
    \begin{algorithmic}[1]
    
        \State $m\_rgb \gets mask(rgb)$ \Comment{Masked RGB}
        \State $m\_depth \gets mask(depth)$ \Comment{Masked depth}
        \State $point\_cloud \gets point\_cloud(m\_rgb, m\_depth, matrix)$
        \State $plane \gets compute\_planes(point\_cloud)$ \Comment{biggest plane}
        \State $normals \gets compute\_normals(plane)$
        \State $center \gets center(plane)$
        \State $orientation \gets halfway\_vector(normal[center])$
        \State $grasp \gets (center, orientation)$
        \\
        \Return $grasp$
    
    \end{algorithmic}
\end{algorithm}

\section{Conclusion and future work}

In this work, we introduced our DoPose-6D dataset, which provides the largest number of samples of bin picking scenes along with samples for tabletop scenes. The dataset contains annotations for different computer vision problems (Instance segmentation, 6D pose estimation, point cloud segmentation, and multi-view methods)

We used the dataset in the context of unseen object segmentation to fine-tune a Mask R-CNN model and showed how the dataset could boost the performance of the core segmentation methods. We also provided insights about using multiple datasets to train unseen object segmentation methods to avoid over-fitting the models.

The grasping pipeline and the pre-trained model are available online as python and ROS packages.

\bibliographystyle{IEEEtran}
\bibliography{IEEEabrv,./references.bib}

\onecolumn

\appendix

\begin{figure}[H]
    \centering
    \begin{tabular}{c c c c}
        Image & ours & UCN \cite{xiang2020learning}  initial & UCN \cite{xiang2020learning}  refined\\
        \includegraphics[width=0.2\textwidth,keepaspectratio]{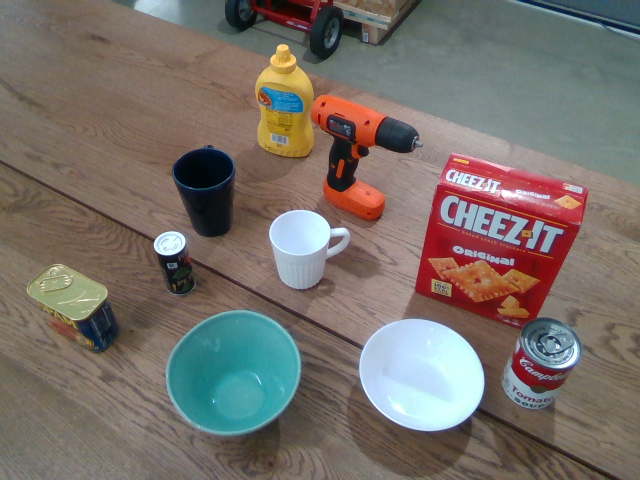}
        &\includegraphics[width=0.2\textwidth,keepaspectratio]{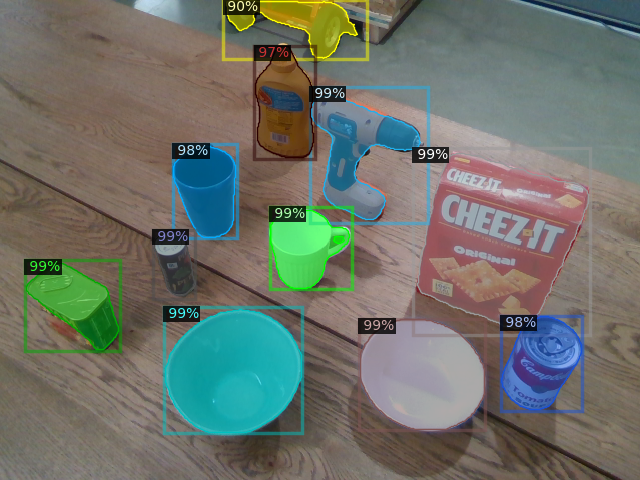}
        &\includegraphics[width=0.2\textwidth,keepaspectratio]{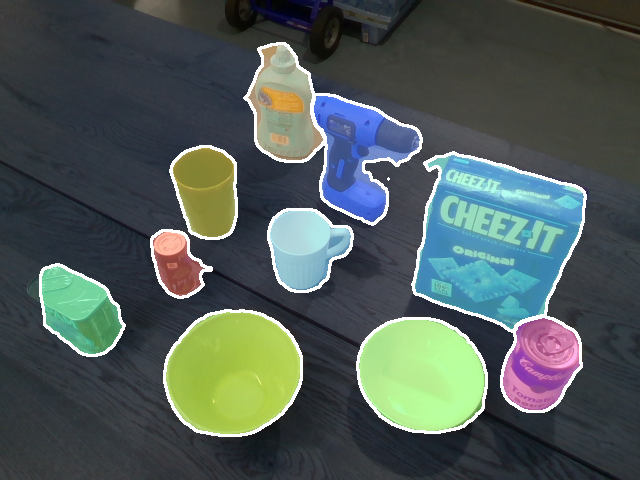}
        &\includegraphics[width=0.2\textwidth,keepaspectratio]{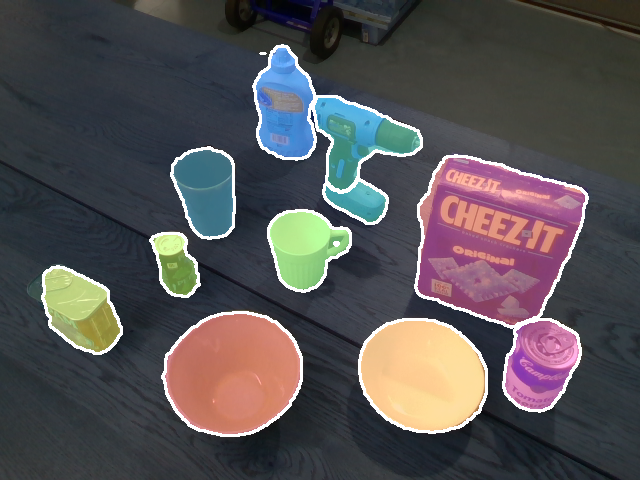}\\
        
        
        \includegraphics[width=0.2\textwidth,keepaspectratio]{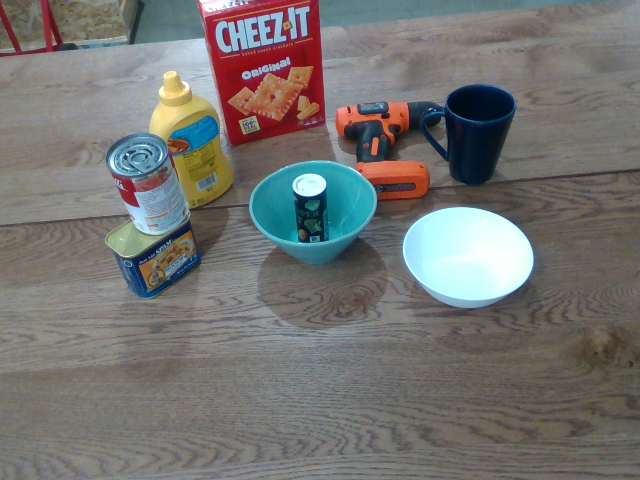}
        &\includegraphics[width=0.2\textwidth,keepaspectratio]{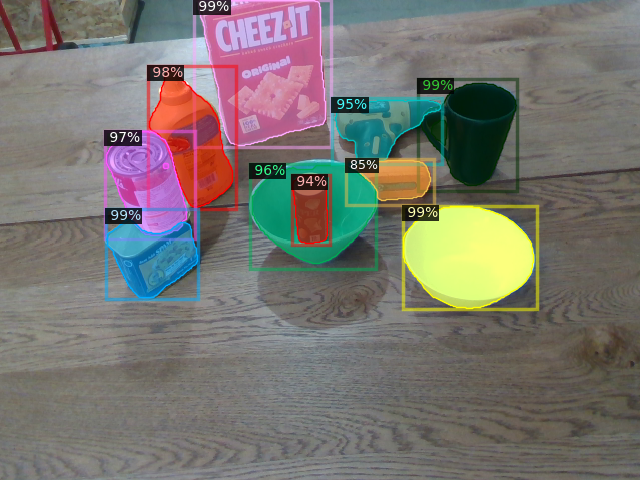}
        &\includegraphics[width=0.2\textwidth,keepaspectratio]{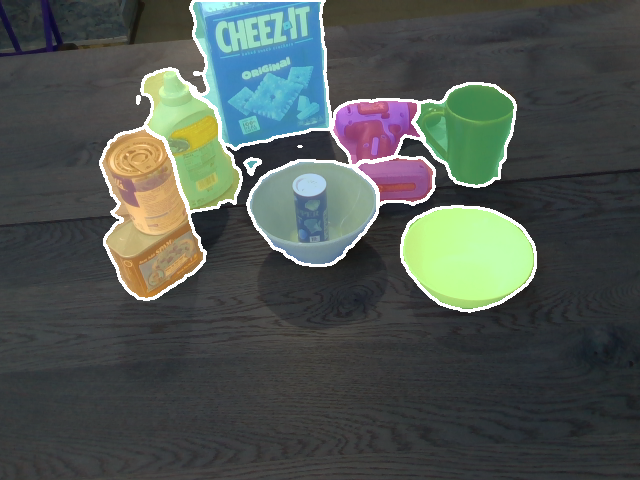}
        &\includegraphics[width=0.2\textwidth,keepaspectratio]{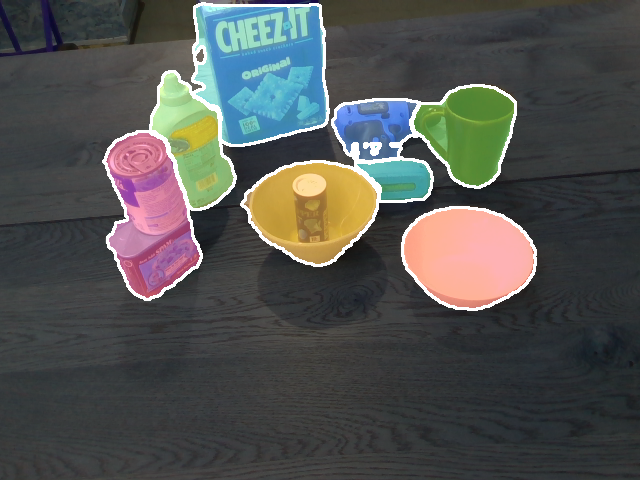}\\
        
        \includegraphics[width=0.2\textwidth,keepaspectratio]{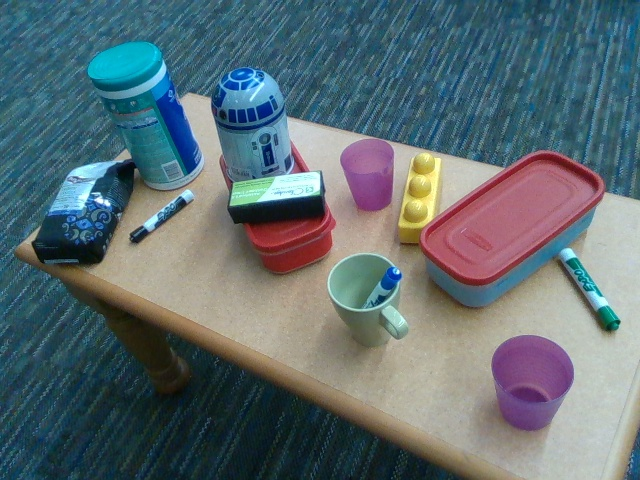}
        &\includegraphics[width=0.2\textwidth,keepaspectratio]{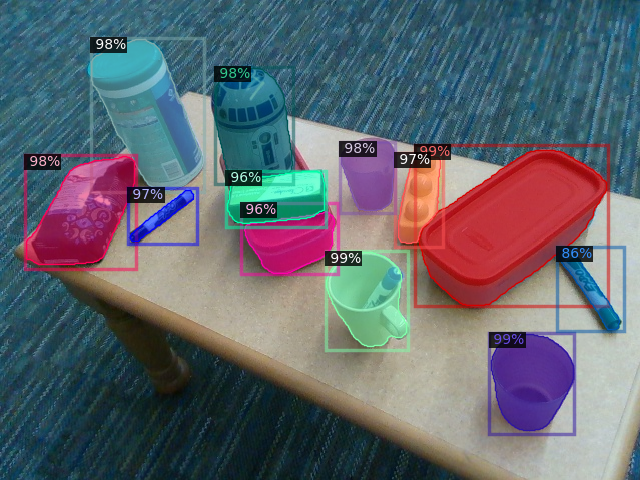}
        &\includegraphics[width=0.2\textwidth,keepaspectratio]{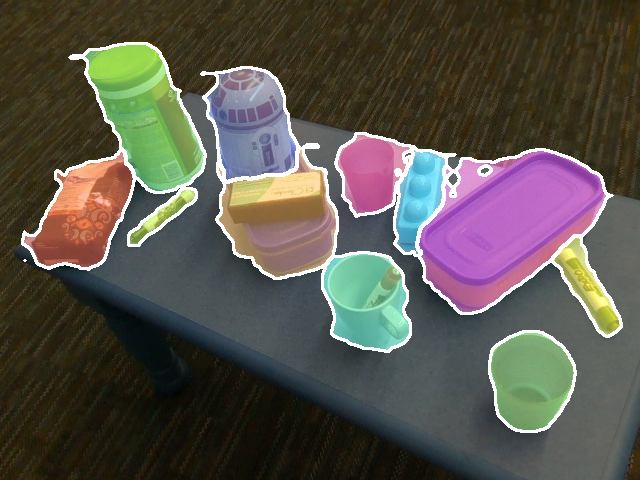}
        &\includegraphics[width=0.2\textwidth,keepaspectratio]{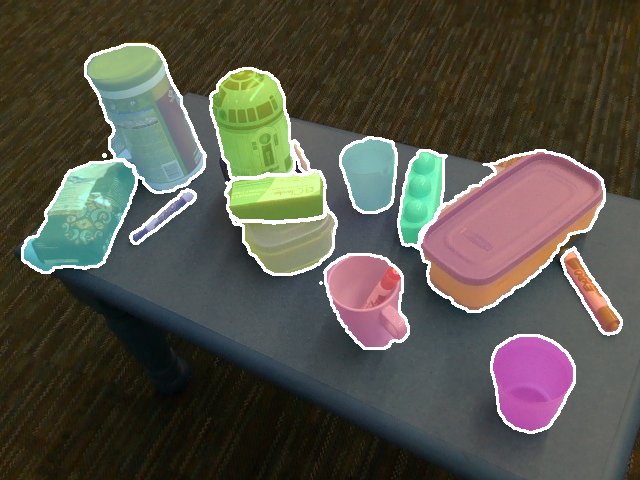}
    \\
    
    \end{tabular}
    \caption{Comparing our model to UCN \cite{xiang2020learning} with its initial and refined segmentation. These example images are provided by the authors in their methods' online repository. We can observe from the figure that our model recognizes clear and sharp object edges against UCN even though our model uses RGB data only. Also, our model could recognize objects placed inside other objects, as shown in the third-row sample.}
    \label{fig:feature_embedding_sota}
\end{figure}

\begin{figure}[H]
    \centering
    \begin{tabular}{c c c c}
        Image & Ground Truth & ours & UOIS \cite{xie2021unseen} \\
        \includegraphics[width=0.2\textwidth,keepaspectratio]{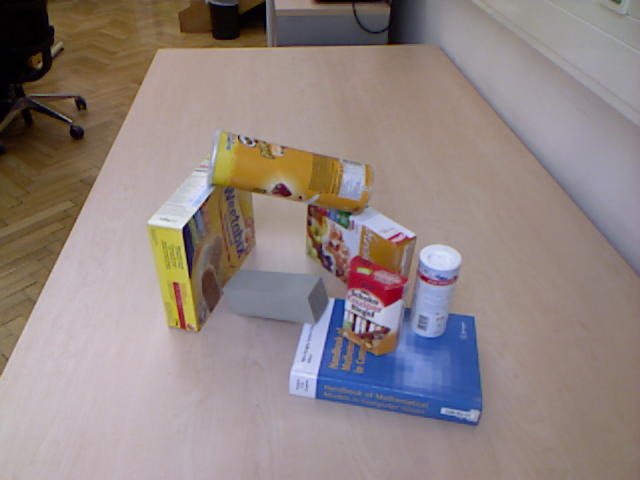}
        &\includegraphics[width=0.2\textwidth,keepaspectratio]{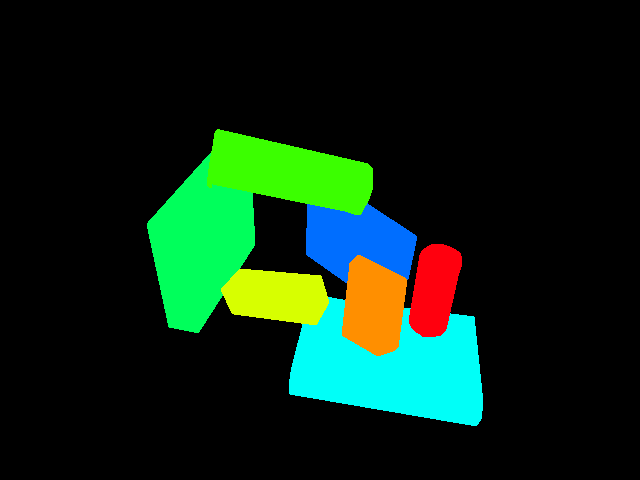}
        &\includegraphics[width=0.2\textwidth,keepaspectratio]{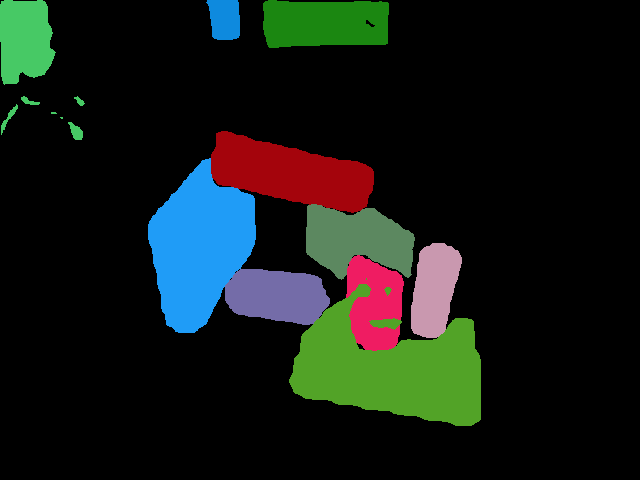}
        &\includegraphics[width=0.2\textwidth,keepaspectratio]{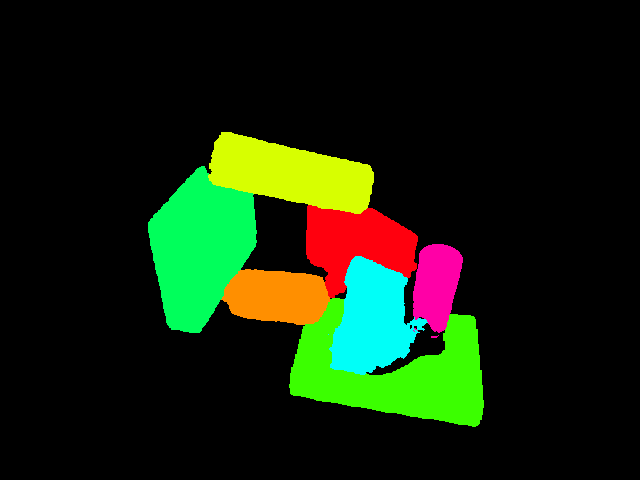}\\
        
        \includegraphics[width=0.2\textwidth,keepaspectratio]{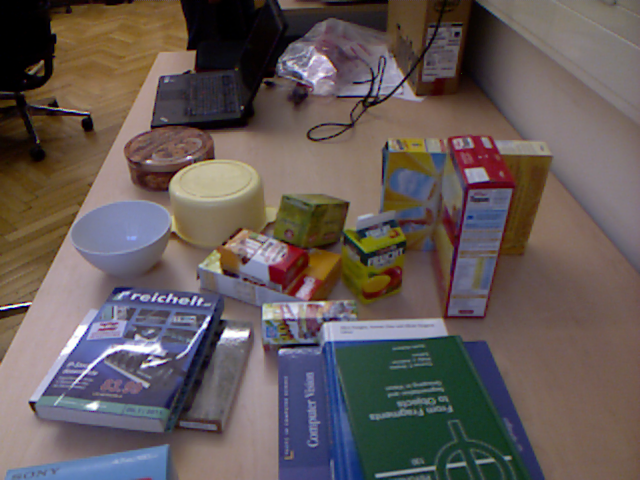}
        &\includegraphics[width=0.2\textwidth,keepaspectratio]{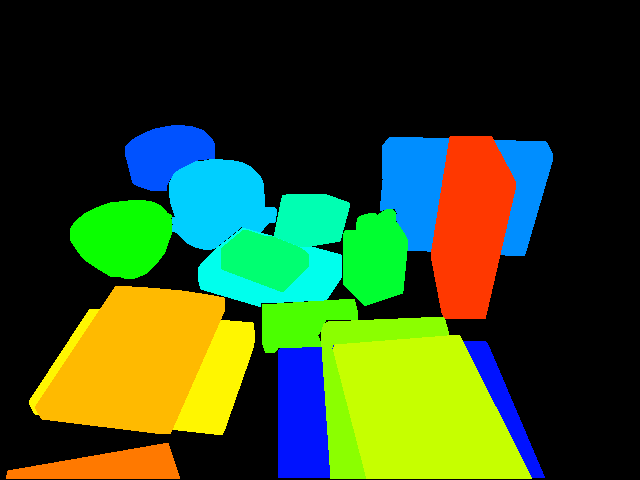}
        &\includegraphics[width=0.2\textwidth,keepaspectratio]{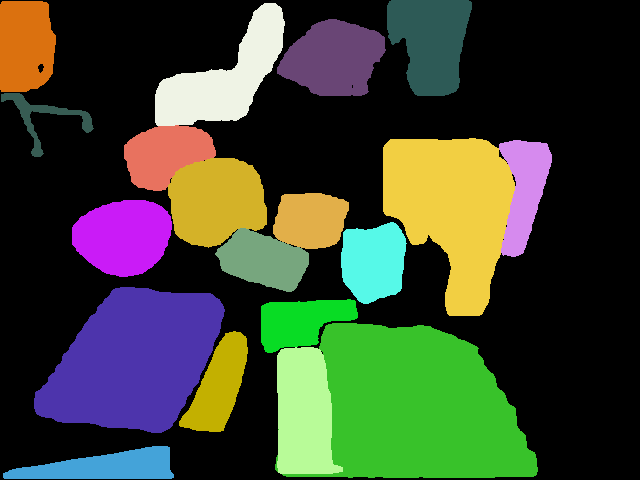}
        &\includegraphics[width=0.2\textwidth,keepaspectratio]{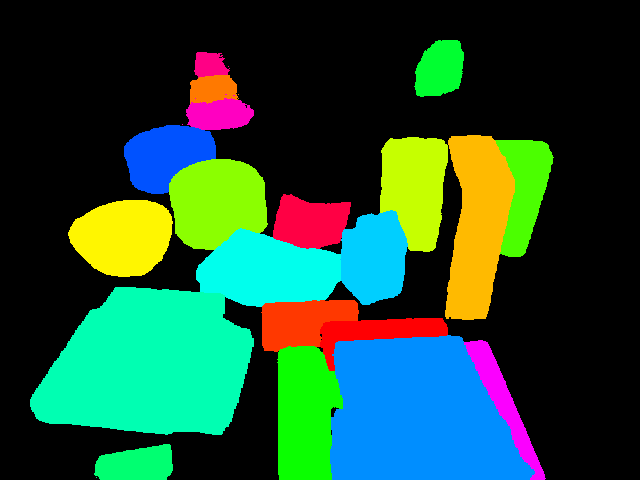}\\

        \includegraphics[width=0.2\textwidth,keepaspectratio]{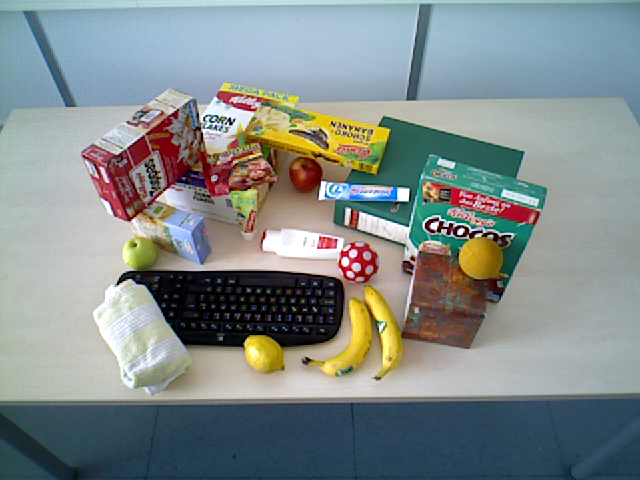}
        &\includegraphics[width=0.2\textwidth,keepaspectratio]{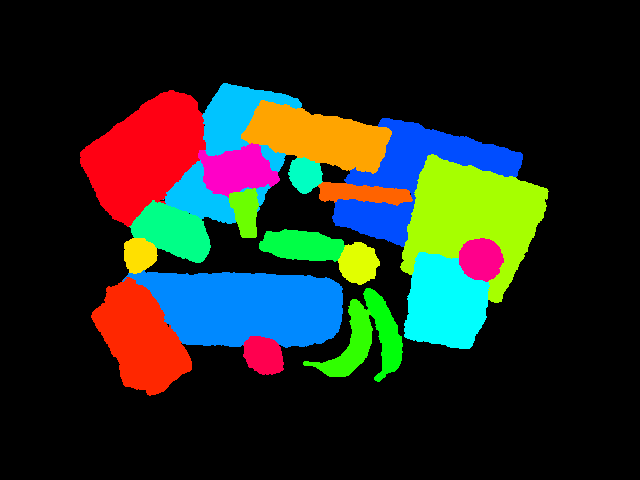}
        &\includegraphics[width=0.2\textwidth,keepaspectratio]{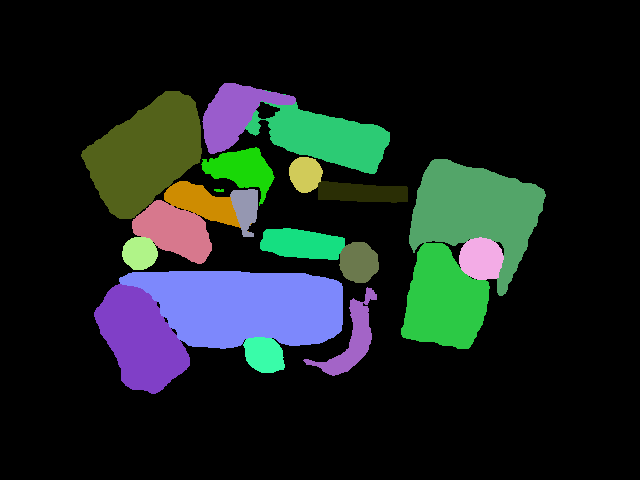}
        &\includegraphics[width=0.2\textwidth,keepaspectratio]{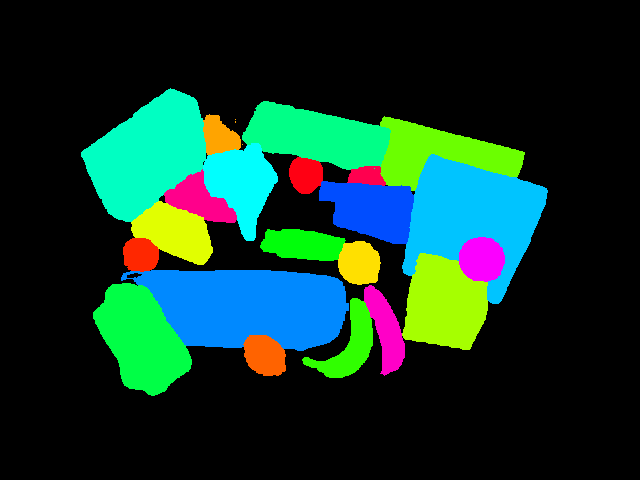}
    \\
    
    \end{tabular}
    \caption{Comparing our model to UOIS \cite{xie2021unseen}. These are samples from the OCID dataset \cite{8793917}. As our model includes no background removal, we notice that some of the background objects are also segmented, but these objects can be excluded with post-processing from the depth image. In general, our model achieved close performance.}
    \label{fig:uois_sota}
\end{figure}

\end{document}